\documentclass{article}
\usepackage{ijcai24}

\usepackage{times}
\usepackage{soul}
\usepackage{url}
\usepackage[hidelinks]{hyperref}
\usepackage[utf8]{inputenc}
\usepackage[small]{caption}
\usepackage{graphicx}
\usepackage{amsmath}
\usepackage{amsthm}
\usepackage{booktabs}
\usepackage{algorithm}
\usepackage{algorithmic}
\usepackage[switch]{lineno}
\usepackage{afterpage}

\nolinenumbers

\usepackage{auxlib}

\title{A Context-Enhanced Framework for Sequential Graph Reasoning}

\makeatletter
\newcommand{\printfnsymbol}[1]{%
  \textsuperscript{\@fnsymbol{#1}}%
}
\makeatother

\author{
Shuo Shi
\and
Chao Peng\thanks{Corresponding authors.} \and
Chenyang Xu\footnotemark[1] \And
Zhengfeng Yang\\
\affiliations
Shanghai Key Laboratory of Trustworthy Computing, Software Engineering Institute, East China Normal University, Shanghai, China\\
\emails
51255902122@stu.ecnu.edu.cn, \\
\{cpeng, cyxu, zfyang\}@sei.ecnu.edu.cn
}

\begin{document}

\maketitle

\begin{abstract}
    


The paper studies sequential reasoning over graph-structured data, which stands as a fundamental task in various trending fields like automated math problem solving and neural graph algorithm learning, attracting a lot of research interest. Simultaneously managing both sequential and graph-structured information in such tasks presents a notable challenge. Over recent years, many neural architectures in the literature have emerged to tackle the issue. In this work, we generalize the existing architectures and propose a context-enhanced framework. The crucial innovation is that the reasoning of each step does not only rely on the outcome of the preceding step but also leverages the aggregation of information from more historical outcomes. The idea stems from our observation that in sequential graph reasoning, each step's outcome has a much stronger inner connection with each other compared to traditional seq-to-seq tasks. We show that the framework can effectively integrate with the existing methods, enhancing their reasoning abilities. Empirical evaluations are conducted on the challenging CLRS Reasoning Benchmark, and the results demonstrate that the proposed framework significantly improves the performance of existing architectures, yielding state-of-the-art results across the majority of the datasets within the benchmark. 

\end{abstract}

\section{Introduction}\label{sec:intro}

Recent years have witnessed a steady growth in exploring the reasoning capacity of deep learning~\cite{DBLP:conf/acl/0001HX023,DBLP:conf/naacl/JiangZWZ22,DBLP:conf/coco/RezendeLN021}. This research domain holds profound significance in the field of artificial intelligence --- it enables AI not only to recognize patterns but also to deduce new knowledge from existing information, thereby broadening the capacity for complex problem-solving. 
As a variety of reasoning tasks can be formulated into graph problems, there is an increasing focus and attention on deep graph reasoning~\cite{DBLP:conf/cvpr/WangY0HP23,DBLP:conf/icdm/LingCJWZCZ22,DBLP:conf/icassp/Zhu22}. This trend delves into the study of graph-structured data and aims to build specialized network structures that can learn to reason over them. 

In this paper, we consider \emph{sequential graph reasoning}, one of the most challenging tasks in deep graph reasoning. Diverging from conventional reasoning tasks like graph (node) classification~\cite{DBLP:conf/kdd/WuHX19}, this type of task goes beyond the one-shot model and targets step-by-step reasoning over graphs. The applications of seq-graph reasoning span across various domains in artificial intelligence. For instance, in AI-powered math problem solving~\cite{DBLP:conf/iclr/LampleC20,DBLP:journals/pr/HuangWLZZYHYW22,DBLP:conf/icmla/DroriKSKMGDDWU20}, neural networks are expected to provide the entire process of solving a mathematical problem, rather than a mere final answer. Another emerging application is neural algorithm learning~\cite{rt,triplet,npq}, where networks are anticipated to sequentially imitate individual steps of classical graph algorithms.

There have been many recent advances in seq-graph reasoning~\cite{DBLP:conf/caibda/LiZZZZ23,DBLP:journals/tim/ZhaoLZZS24,DBLP:journals/access/GerasimovaSM23}. 
While the proposed neural architectures exhibit diversity, they generally adhere to a common sequential paradigm. In the framework, graph-structured hidden states are maintained in memory. At each step, the outcome of the preceding step serves as input, integrating with the hidden states for reasoning. Variations among existing works arise in the specifics of their integration processes. Numerous graph networks (e.g. GAT~\cite{gat} and MPNN~\cite{mpnn}) that excel at one-shot reasoning have been investigated.






The current paradigm is essentially a direct combination of the techniques for processing serialized data and those for handling graph-structured data. We notice that such a straightforward combination may fall short of yielding competitive reasoning capabilities due to the distinctive nature of seq-graph reasoning. Compared to traditional seq-to-seq tasks~\cite{DBLP:phd/ethos/Yu17b}, each step's outcome in seq-graph reasoning has a much stronger intrinsic connection with the others. For example, in automatic geometric
problem solving~\cite{DBLP:journals/mics/SchreckM16}, the outcome sequence forms an entire problem-solving process, which can even be represented by an extra reasoning graph. However, most existing works fail to capture these intrinsic connections.



\subsection{Our Contributions}

Motivated by the observation above, this paper proposes a framework to capture the contextual patterns of step outcomes in seq-graph reasoning. To evaluate the reasoning ability of this framework, we conduct empirical evaluations on the challenging CLRS Reasoning Benchmark~\cite{clrs}. This benchmark consists of 30 different algorithmic tasks for probing the reasoning capabilities of graph-structured networks. The contributions of the paper are summarized as follows:

\begin{itemize}
    \item We introduce a \emph{Context-Enhanced} Framework (CEF) for seq-graph reasoning. The framework extends the existing paradigm and exhibits notable flexibility, allowing for adjustments based on available memory space during implementation.
    \item We show that the framework can effectively integrate with various existing architectures. Specifically, the paper categorizes existing methods into two types: GNN-based and Transformer-based, and provides detailed CEF implementations for both.
    \item We perform a diverse range of experiments on the CLRS benchmark. The results demonstrate that, for both types of existing architectures, our framework significantly enhances their performances across the majority of the datasets within the benchmark. When integrated with the latest architectures, we achieve new state-of-the-art results.
\end{itemize}

\subsection{Related Work}\label{subsec:related}

\paragraph{Graph Neural Networks.} Graph neural networks have emerged as a powerful tool for effectively capturing and leveraging the structural relationships present in graph-structured data. Numerous neural architectures have been proposed in this domain, such as Graph Convolution Networks (GCN)~\cite{gcn}, Graph Attention Networks (GAT)~\cite{gat}, Message Passing Neural Networks (MPNNs)~\cite{mpnn} and so on. 
These architectures excel not only in one-shot graph reasoning~\cite{DBLP:journals/datamine/MaMMZL023} but also demonstrate comparable performance in seq-graph reasoning tasks~\cite{DBLP:conf/acl/RameshSH23}. Recently,~\cite{triplet} introduced a novel GNN architecture called triplet-GMPNN. They augment the classical MPNN to perform message passing toward edges and have achieved remarkable results in neural algorithmic reasoning.

\paragraph{Graph Transformers.} Transformers and its variants play a crucial role in domains such as natural language processing, computer vision, and time series analysis~\cite{transformer}. Consequently, leveraging Transformers for tasks related to graph reasoning has emerged as a recent research focus, leading to the development of several variants of Transformers tailored for handling graph-structured data~\cite{trans1,hussain2022global}. A recent advancement in this direction is proposed by~\cite{rt}. 
They introduced Relational Transformers, which enhance attention computation by aggregating edge and graph feature information with node features. 

\paragraph{Beyond One-Shot Graph Reasoning.} The architectures mentioned above are fundamentally tailored for one-shot reasoning. When addressing seq-graph reasoning tasks, they necessitate the paradigm stated in the introduction. There have been some recent advances in exploring competitive paradigms for seq-graph reasoning~\cite{dual,npq,recursive_ar}. Notably, the studies most closely related to our research were recently introduced by~\cite{npq} and~\cite{recursive_ar}, where the GNN architectures are augmented with a priority queue and a stack, respectively. 

\subsection{Paper Organization}

The paper is organized as follows.~\cref{sec:pre} provides preliminaries, including a problem definition and a paradigm widely used in the literature. In~\cref{sec:framework}, we introduce our framework, and subsequently,~\cref{sec:implement} details the concrete implementations of our framework for different types of existing approaches. Empirical evaluations are presented in~\cref{sec:experiment}. Finally,~\cref{sec:conclusion} concludes the paper.  
\section{Preliminaries}\label{sec:pre}

The section formalizes the problem setup and introduces a widely-used paradigm. In a seq-graph reasoning task, we are given graph-structured data and required to perform reasoning over it at most $T$ times. 
The given data is represented by a graph $G=(V, E)$ with $n$ nodes and $m$ edges, where each node $v\in V$ has feature $x^{(0)}_v$ and each edge $e\in E$ has feature $x^{(0)}_e$. For the notational simplicity, define $X^{(0)}_V:= \{x^{(0)}_v\}_{v\in V}$, $X^{(0)}_E:= \{x^{(0)}_e\}_{e\in E}$, and $\vX^{(0)} = \{ X^{(0)}_V, X^{(0)}_E \}$.
Further, for each training data, we are provided with the ground truth $\vY^{(t)}$ of each reasoning step $t\in [T]$. 

A widely used paradigm in the literature is the \emph{encode-process-decode} paradigm~\cite{triplet}. In this paradigm, there are three modules: encoder $\cE$, processor $\cP$, and decoder $\cD$. In each step $t\in [T]$, the input tensors $\vX^{(t-1)}$ undergo three successive transformations $\cE, \cP, \cD$ and then yield $ \vX^{(t)} $, which serves as the input for the next step. 

\paragraph{Encoder Module.} There are usually two encoder networks in this module\footnote{Some prior works add one extra encoder network to handle the global feature of the graph.}, each tasked with encoding node features and edge features respectively. Thus, after feeding $\vX^{(t-1)}$ to the encoder module, we obtain latent features $\vL^{(t)}$ for nodes and edges in the graph:
\begin{equation}\label{eq:encoder}
    L^{(t)}_{V}, L^{(t)}_{E} = \cE\left(\vX^{(t-1)}\right).
\end{equation}

\begin{figure*}[tb]
    \centering
    \includegraphics[width=1.0\textwidth]{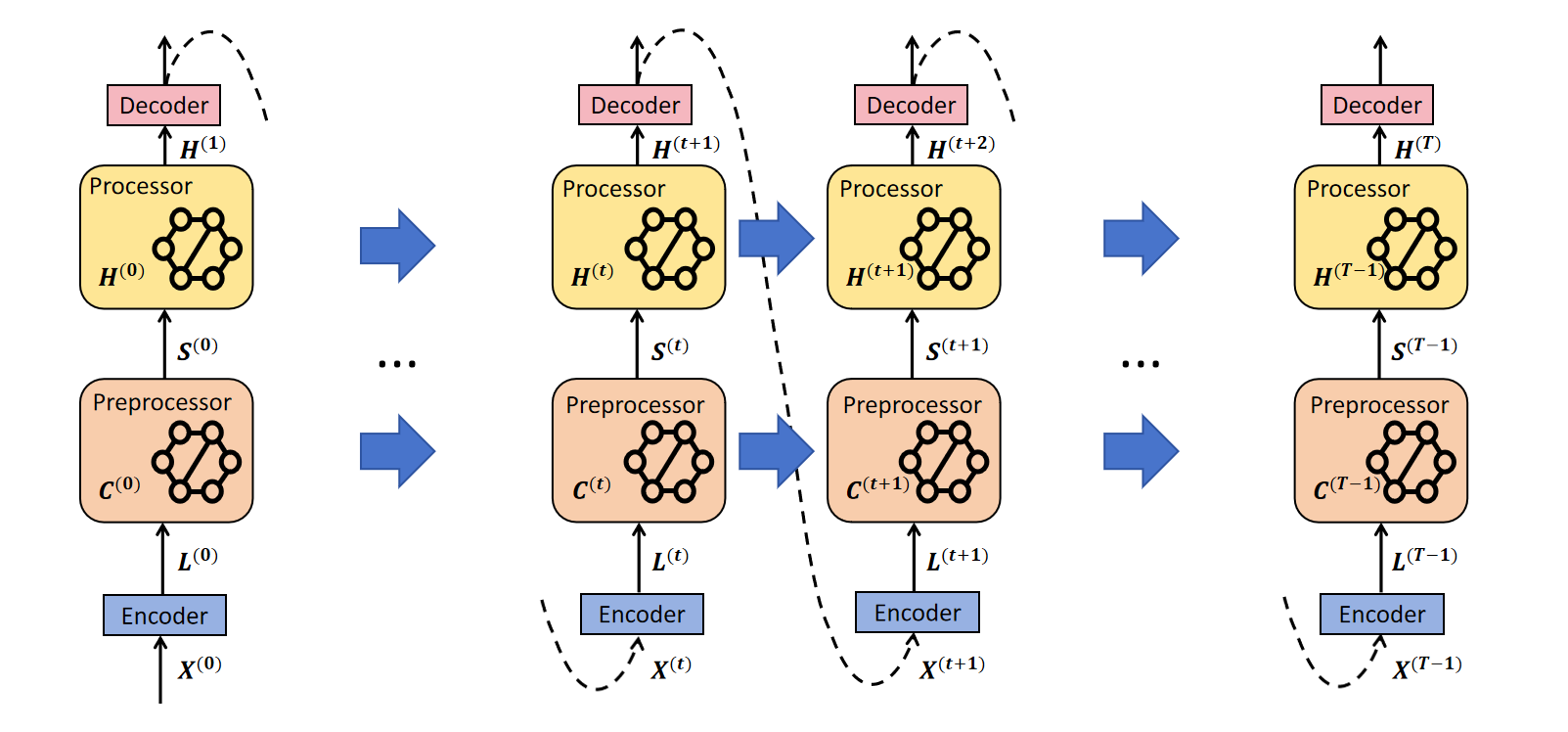}
    \caption{An illustration of the CEF framework. The solid and dashed arrows in the figure indicate the direction of data flow for the graph structure during the inference process. The blue arrows represent the historical information aggregation operation performed within the framework.}
    \label{fig:cef}
\end{figure*}

\paragraph{Processor Module.} This module is the central component of the entire paradigm, where most prior works primarily vary. It consists of a (standard) graph-structured network (e.g., graph neural networks or graph transformers), and maintains a hidden state for each node and edge. Denote by $H^{(t-1)}_V, H^{(t-1)}_E$ the hidden states of the node set and the edge set at the beginning of step $t$ respectively. First, we concatenate the latent feature and the hidden state of each node and edge, obtaining their new representations:
\begin{equation}\label{eq:pro_concate}
    Z^{(t)}_V = \left.L^{(t)}_{V} \;\middle\Vert\; H^{(t-1)}_V \right. , \;\;\;\;\; Z^{(t)}_E = \left. L^{(t)}_{E} \;\middle\Vert\; H^{(t-1)}_E  \right. .
\end{equation}
And then, feed these representations into the graph-structured network, obtaining new hidden states:
\begin{equation}\label{eq:pro}
    H^{(t)}_V, H^{(t)}_E = \cP\left(Z^{(t)}_V, Z^{(t)}_E \right).
\end{equation}

\paragraph{Decoder Module.} Without loss of generality, we can assume that this module contains two networks to decode the hidden states of nodes and edges respectively:
\begin{equation}\label{eq:decoder}
    X^{(t)}_V, X^{(t)}_E = \cD\left(H^{(t)}_V, H^{(t)}_E \right).
\end{equation}
Given the ground truth $\vY^{(t)}$ of each step, we can compute the total loss and thus, train the entire network.



\section{CEF: A Context-Enhanced Framework}\label{sec:framework}

The section presents a framework that generalizes the standard encode-process-decode paradigm. Before stating the details, let us give some intuition first. Compared to the traditional seq-to-seq tasks, we observe that the ground truths in seq-graph reasoning have a much stronger inner connection with each other. For example, in neural graph algorithm learning, the goal is to imitate the execution of a classical graph algorithm like Prim's algorithm~\cite{6773228}.
In this scenario, each $\vY^{(t)}$ corresponds to an individual step of Prim's algorithm, and the entire ground truth sequence represents the process of solving the minimum spanning tree problem by Prim's algorithm, which is an optimization problem-solving procedure. Thus, similar to the consideration of momentum in historical gradients during gradient descent, in seq-graph reasoning, momentum concerning the outcome at each step should also be introduced and taken into account. This observation motivates our framework.

The framework builds upon the encode-process-decode paradigm and newly introduces a \emph{Preprocessor} module between the encoder and processor. See~\cref{fig:cef} for an illustration. This module is designed to capture dependencies and patterns on the latent feature sequence $<\vL^{(1)},\vL^{(2)},...,\vL^{(t)},...>$, which are also referred to as ``context'' information. Hence, we term the new framework a \emph{Context-Enhanced Framework} (CEF).  
More precisely, the module maintains a context state for each node and edge, storing the information aggregated from historical latent features. For each node $v$ (resp. edge $e$), denote by $c^{(t)}_v$ (resp. $c^{(t)}_e$) the context state at the beginning of step $t$. Note that $c^{(t)}_v$ can be a direct concatenation of historical latent features or the outcome of applying some dimensionality reduction techniques (e.g. gating mechanisms) to them. For simplicity, $ C^{(t-1)}_V $, $C^{(t-1)}_E $ and $\vC^{(t-1)}$ are defined similarly. 

The module involves two pivotal operations: context enhance function $\vf_{\enhance}$ and context update function $\vf_{\update}$. 
The application of $\vf_{\enhance}$ aims to augment latent features with contextual information, subsequently serving as input to the processor module:
\begin{equation}\label{eq:enhance}
    \vS^{(t)} = \vf_{\enhance}\left( \vL^{(t)}, \vC^{(t)} \right)
\end{equation}
Concurrently, $\vf_{\update}$ is employed to update the context state of each node and edge:
\begin{equation}\label{eq:update}
    \vC^{(t+1)} = \vf_{\update}\left( \vL^{(t)}, \vC^{(t)} \right)
\end{equation}


The most critical aspect within the framework is how to set appropriate $\vf_{\update}$ and $\vf_{\enhance}$ so that they can integrate well with various types of processor modules. 
We notice that a popular approach to capturing patterns in historical latent features is through the use of the attention mechanism. Specifically, we can employ a node-wise (resp. edge-wise) concatenation function for $\vf_{\update}$ and a QKV attention network for $\vf_{\enhance}$. However, this method proves to be quite resource-intensive. The demand for memory significantly escalates with an increase in the total number of reasoning steps. An experimental study on this implementation is provided in the paper's full version. In the main body, our focus shifts to leveraging \emph{gating mechanisms} in the preprocessor module --- a more space-efficient approach that demonstrates comparable empirical performance.

\section{Applications}\label{sec:implement}

As outlined in~\cref{subsec:related}, existing (processor) architectures fall within two categories: GNN-based and Transformer-based. This section considers both processor types and presents specific implementations of the framework for each. 


\subsection{Integration with GNN-based Processors}\label{subsec:GNN}

This subsection considers GNN-based processors. We follow the literature~\cite{triplet} and formulate this type of processor into the model below. Note that the existing GNN-based processors only maintain hidden states for nodes. 

In each step $t\in [T]$,
\setlength{\FrameSep}{2pt}
\begin{framed}
\begin{equation}
\begin{aligned}
\forall v\in V: z_v^{(t)}&=\left. s_v^{(t)} \;\middle\Vert\; h_v^{(t-1)}\right. \\
r_v^{(t)} &= \vf_1\left(z_v^{(t)}\right) \\
    m_{v} &= \bigoplus\limits_{u:(u,v)\in E} \vf_2\left(r_v^{(t)},r_u^{(t)}, s_{(u,v)}^{(t)}  \right) \\
    h_v^{(t)} &= \vf_3\left(r_v^{(t)}, m_{v}\right)
\end{aligned}
\end{equation}
\end{framed}

Various processors may utilize distinct $\vf_1$, $\vf_2$, $\vf_3$ and aggregation function $\bigoplus$. Since only node hidden states are preserved and computed in such processors, we also focus on the transformation of node latent features in the preprocessor module and let $s_e^{(t)} = \ell_e^{(t)}$ directly for each edge $e\in E$. We introduce a novel gating mechanism to preprocess each node latent feature. Further, the weight sharing technique~\cite{DBLP:journals/taco/GarlandG18} is employed to make the module more memory-efficient and reduce the risk of overfitting. We implement the preprocessor as follows:

\setlength{\FrameSep}{2pt}
\begin{framed}
\begin{equation}
\begin{aligned}
   \forall v\in V: \alpha_v &=  \relu\left( \otanh \left( \vf_{\linear}\left(c_v^{(t)}\right) \right) \right) \\
    s_v^{(t)} & = c_v^{(t+1)}  = \alpha_v \cdot c_v^{(t)} + (1-\alpha_v) \cdot \ell_v^{(t)}\\
\end{aligned}
\end{equation}
\end{framed}

$\vf_{\linear}$ is a linear layer that transforms $c_v^{(t)}$ into a one-dimensional scalar. Then, we apply function $\otanh$ and $\relu$ sequentially, yielding $\alpha_v\in [0,1)$. Interpreting $\alpha_v$ as a forget factor, $s_v^{(t)}$ and $c_v^{(t+1)}$ are obtained by a linear combination of the context state and latent feature. We provide an experimental study and discussion (in the full version of the paper) to explain why we chose the $\sigmoid$ function. In summary, empirical findings suggest that when integrating with GNN processors, the latent features in seq-graph reasoning exhibit distinctive patterns and prefer a higher probability of the forget factor being $0$.



\subsection{Integration with Transformer-based Processors}

This subsection considers Transformer-based processors. They maintain hidden states for both nodes and edges.  We propose a model below to capture the common structures of such processors. 
Use $\cN(v):=\left\{u|(v,u)\in E\right\}$ to denote the neighbors of $v$. In each step $t\in [T]$,  

\setlength{\FrameSep}{2pt}
\begin{framed}
\begin{align}
    \forall v\in V:\;\; & \nonumber\\
q_v &= \vf_{\query}\left(z_v^{(t)},\; \left\{ z_{(v,u)}^{(t)}\right\}_{u\in \cN(v)} \right) \nonumber\\ 
k_v & = \vf_{\key}\left(\left\{z_u^{(t)}, \; z_{(v,u)}^{(t)}\right\}_{u\in \cN(v)}\right) \nonumber\\
\vartheta_v & = \vf_{\ovalue}\left( \left\{z_u^{(t)}, \; z_{(v,u)}^{(t)}\right\}_{u\in \cN(v)} \right) \\
h_v^{(t)} & = \vf_{\node}\left(z_v^{(t)},q_v, k_v, \vartheta_v \right) \nonumber\\
\forall (u,v)\in &E: \nonumber\\
h_{(u,v)}^{(t)}&= \vf_{\edge} \left( h_u^{(t)},h_v^{(t)}, z_{(u,v)}^{(t)}, z_{(v,u)}^{(t)}   \right)\nonumber
\end{align}
\end{framed}

The model extends the QKV attention mechanism, taking into account the incorporation and subsequent updating of edge features. Different processors employ different $\vf$ functions within the model. 
Certainly, we could adopt a context enhancement approach similar to the one in~\cref{subsec:GNN}. However, we notice that, unlike existing GNN processors, the Transformer processors in the literature do not utilize gating mechanisms in the update of hidden states. Thus, to capture the contextual information of latent features and hidden states simultaneously, 
we propose a refined framework that seamlessly integrates the preprocessor and processor, allowing for mutual sharing of states.  Initially, the preprocessor utilizes hidden states for context state updates. 
\setlength{\FrameSep}{2pt}
\begin{framed}
\begin{align}
    \forall v\in V:\;\;\;\;\; & \nonumber\\
z_v^{(t)} &= \left. \ell_v^{(t)} \;\middle\Vert\; h_v^{(t-1)} \right. \nonumber\\
\alpha_v &= \sigmoid\left(\vf_{\linear\textendash 1} \left(c_v^{(t)}\right) \right) \nonumber\\
c_v^{(t+1)} & = \alpha_v \cdot c_v^{(t)} + (1-\alpha_v) \cdot z_v^{(t)} \nonumber\\
\forall e\in E: \;\;\;\;\; & \\
z_e^{(t)} &= \left. \ell_e^{(t)} \;\middle\Vert\; h_e^{(t-1)} \right. \nonumber\\
\alpha_e &= \sigmoid\left(\vf_{\linear\textendash 2} \left(c_e^{(t)}\right) \right) \nonumber\\
c_e^{(t+1)} & = \alpha_e \cdot c_e^{(t)} + (1-\alpha_e) \cdot z_e^{(t)} \nonumber
\end{align}
\end{framed}
Subsequently, the processor employs the resulting context states to update hidden states. 

\setlength{\FrameSep}{2pt}
\begin{framed}
\begin{align}
\forall v\in V:\;\;\;\;\; & \nonumber\\
q_v &= \vf_{\query}\left(z_v^{(t)},\; \left\{ z_{(v,u)}^{(t)}\right\}_{u\in \cN(v)} \right) \nonumber\\ 
k_v & = \vf_{\key}\left(\left\{c_u^{(t+1)}, \; c_{(v,u)}^{(t+1)}\right\}_{u\in \cN(v)}\right) \nonumber\\
\vartheta_v & = \vf_{\ovalue}\left( \left\{c_u^{(t+1)}, \; c_{(v,u)}^{(t+1)}\right\}_{u\in \cN(v)} \right) \\
h_v^{(t)} & = \vf_{\node}\left(z_v^{(t)},q_v, k_v, \vartheta_v\right) \nonumber\\
\forall (u,v)\in E: & \nonumber\\
h_{(u,v)}^{(t)}&= \vf_{\edge} \left( h_u^{(t)},h_v^{(t)}, c_{(u,v)}^{(t+1)}, c_{(v,u)}^{(t+1)}   \right) \nonumber
\end{align}
\end{framed}

$\vf_{\linear\textendash 1}$ and $\vf_{\linear\textendash 2}$ are two linear layers that transform the context states of nodes and edges into one-dimensional scalars, respectively. Then we apply function $\sigmoid$, yielding forget factors and updating context states. 
This process enables the context states to capture contextual information from both latent features and hidden states.

Next, we compute $q_v,k_v,\vartheta_v$. 
While preserving the original computation for $q_v$, we strategically substitute $\vZ^{(t)}$ with the context states $\vC^{(t+1)}$ during the computation of $k_v$ and $\vartheta_v$.
In essence, departing from conventional self-attention, we empower each node to aggregate attention information from the context states. Finally, we leverage $q_v,k_v,\vartheta_v$ to update the hidden state for each node and edge. 




\section{Experiments}\label{sec:experiment}

This section validates the empirical performance of our framework and aims to investigate the following questions:
\begin{itemize}
    \item Can the framework simultaneously enhance the reasoning capabilities of GNN-based and Transformer-based architectures? If so, is there a noticeable discrepancy in the performance improvements between these two architectural types?
    \item Consider an architecture faced with a reasoning task in which it inherently performs sub-optimally. If our framework is applied, can we expect an enhancement in performance? In simpler terms, does the framework push the original architecture towards a more ``extreme'' or improved state?
\end{itemize}

To this end, we apply the framework to two state-of-the-art architectures: Triplet-GMPNN~\cite{triplet}, a GNN-based model, and RT~\cite{rt}, a Transformer-based model. We refer to their implementations as CEF-GMPNN and CEF-RT, respectively. We remark that in the main body, we focus on the exploration of the aforementioned two questions; the ablation experiments regarding implementation details are deferred to the appendix. The code is available at \href{https://github.com/Ghost-st/CEF}{https://github.com/Ghost-st/CEF}.


\subsection{Setup}


The experiments are conducted on a machine equipped with an i7-13700K CPU and an RTX 4090 GPU. The results are averaged over 5 runs with random seeds 5, 18, 25, 30, and 42.

\paragraph{Datasets.} We use the CLRS Algorithmic Reasoning Benchmark~\cite{clrs}, a proven benchmark that offers a unified evaluation (micro-F1 score) for assessing the (seq-graph) reasoning capabilities of neural networks. It is derived from a foundational algorithms textbook ``Introduction to Algorithms''~\cite{cormen2022introduction} and consists of 30 algorithmic reasoning tasks, covering a wide range of algorithmic procedures including sorting, searching, dynamic programming, graph algorithms, string algorithms, and geometric algorithms. The inclusion of such a variety of tasks enables us to draw meaningful conclusions about the effectiveness of the proposed framework.

\paragraph{Baselines.} The two most crucial baselines we compare against are Triplet-GMPNN~\cite{triplet} and RT~\cite{rt}, which play a pivotal role in addressing the aforementioned questions. 
Moreover, we include Memnet~\cite{memnet}, MPNN~\cite{mpnn}, PGN~\cite{pgn}, and NPQ~\cite{npq} in experiments as important references for comparison.


\paragraph{Training Details.} In our implementations, $\vf_{\linear}$, $\vf_{\linear\textendash 1}$ and $\vf_{\linear\textendash 2}$ are individual linear layers.
To ensure fair comparisons, we employ the experimental hyperparameter settings used in triplet-GMPNN~\cite{triplet} and Relational Transformer~\cite{rt} for CEF-GMPNN and CEF-RT, respectively. Specifically, for CEF-GMPNN, we set the batch size to 32 and the network is trained for 10,000 steps by Adam optimizer with a learning rate of 0.001; 
 while for CEF-RT, we set the batch size to 4 and the network is trained for 10,000 steps by Adam optimizer with a learning rate of 0.00025.  We remark that with the introduction of a preprocessor module in our framework, the average training time increases by 3 to 5 minutes for both methods. This increment represents less than 10\% of the total runtime, indicating the efficiency of our implementations. See~\cref{sec:time} for more details.

\begin{figure*}[tb]
    \centering
    \includegraphics[width=1.0\textwidth]{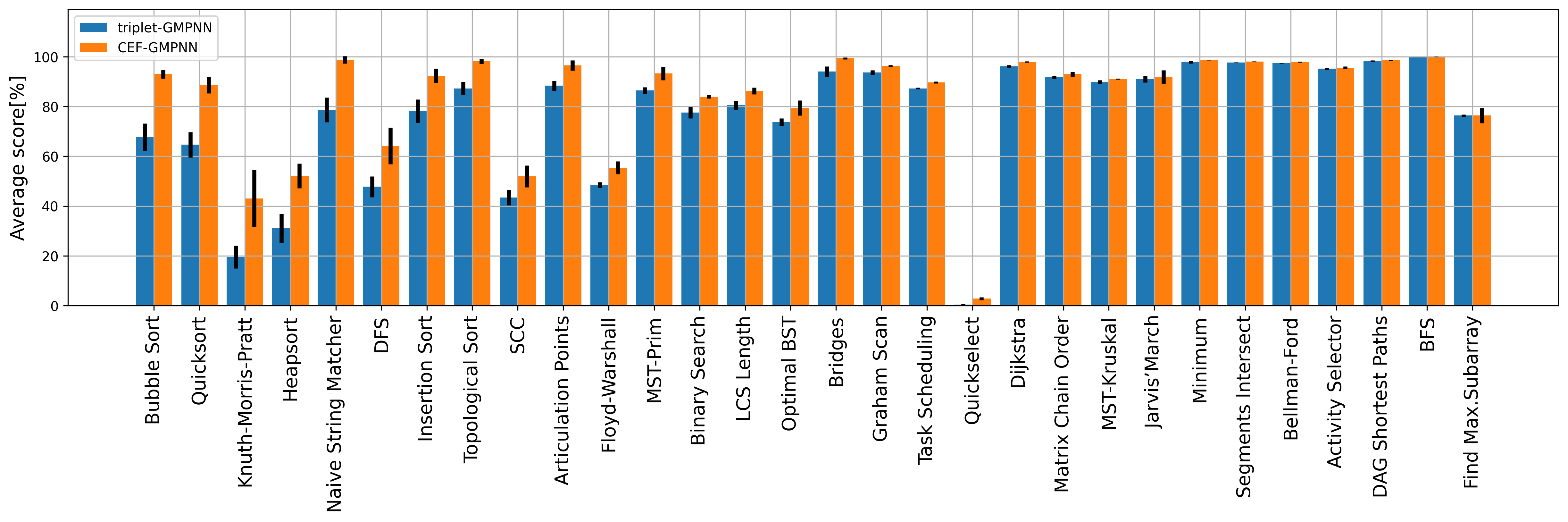}
    \caption{Comparisons between our CEF-GMPNN and Triplet-GMPNN. The 30 tasks are arranged in descending order of improvement magnitude.}
    \label{fig:mpnn}
\end{figure*}

\begin{figure*}[tb]
    \centering
    \includegraphics[width=1.0\textwidth]{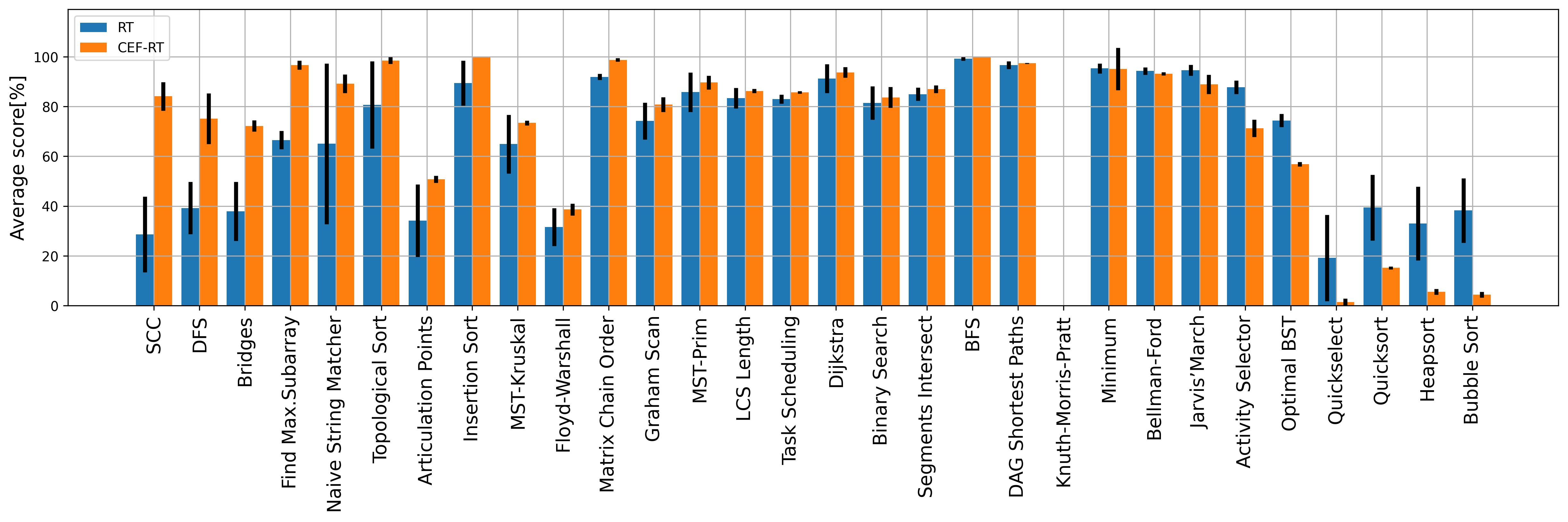}
    \caption{Comparisons between our CEF-RT and RT. The 30 tasks are arranged in descending order of improvement magnitude. Note that on the Knuth-Morris-Pratt reasoning task, both algorithms exhibit very poor performance. Refer to~\cref{tab:results} for detailed scores.}
    \label{fig:rt}
\end{figure*}

\begin{table*}[tb]
    \centering
    \begin{tabular}{cccccc}
    \hline
        Tasks & Prior Best & Triplet-GMPNN & RT & CEF-RT (ours) & CEF-GMPNN (ours) \\ \hline
        Activity Selector & 83.36\%±4.27  & 95.18\%±0.45 & 87.72\%±2.7 & 71.22\%±3.51 & \textbf{95.58\%±0.49}  \\ 
        Articulation Points & 50.91\%±2.18  & 88.32\%±2.01 & 34.15\%±14.6 & 50.71\%±1.42 & \textbf{96.51\%±2.09}  \\ 
        Bellman-Ford & 92.99\%±0.34  & 97.39\%±0.19 & 94.24\%±1.5 & 93.15\%±0.68 & \textbf{97.80\%±0.20}  \\ 
        BFS & 99.89\%±0.05  & 99.73\%±0.04 & 99.14\%±0.7 & \textbf{99.99\%±0.00} & 99.95\%±0.09  \\ 
        Binary Search & 76.95\%±0.13  & 77.58\%±2.35 & 81.48\%±6.7 & 83.66\%±4.18 & \textbf{83.91\%±0.70}  \\ 
        Bridges & 72.69\%±4.78  & 93.99\%±2.07 & 37.88\%±11.8 & 72.21\%±2.24 & \textbf{99.36\%±0.38}  \\ 
        Bubble Sort & 73.58\%±0.78  & 67.68\%±5.50 & 38.22\%±13.0 & 4.35\%±1.16 & \textbf{92.97\%±1.73}  \\
        DAG Shortest Paths & 96.94\%±0.16  & 98.19\%±0.30 & 96.61\%±1.6 & 97.35\%±0.22 & \textbf{98.55\%±0.19}  \\ 
        DFS & 13.36\%±1.61  & 47.79\%±4.19 & 39.23\%±10.5 & \textbf{75.11\%±10.23} & 64.17\%±7.33  \\ 
        Dijkstra & 91.50\%±0.50  & 96.05\%±0.60 & 91.2\%±5.8 & 93.69\%±2.14 & \textbf{97.89\%±0.22}  \\ 
        Find Max.Subarray & 65.23\%±2.56  & 76.36\%±0.43 & 66.52\%±3.7 & \textbf{96.63\%±1.77} & 76.38\%±3.04  \\ 
        Floyd-Warshall & 28.76\%±0.51  & 48.52\%±1.04 & 31.59\%±7.6 & 38.61\%±2.42 & \textbf{55.42\%±2.58}  \\ 
        Graham Scan & 91.04\%±0.31  & 93.62\%±0.91 & 74.15\%±7.4 & 80.74\%±2.97 & \textbf{96.27\%±0.37}  \\ 
        Heapsort & \textbf{68.00\%±1.57}  & 31.04\%±5.82 & 32.96\%±14.8 & 5.57\%±1.17 & 52.13\%±4.91  \\ 
        Insertion Sort & 71.42\%±0.86  & 78.14\%±4.64 & 89.43\%±9.0 & \textbf{99.99\%±0.00} & 92.41\%±2.84  \\ 
        Jarvis’March & 92.88\%±2.87  & 91.01\%±1.30 & \textbf{94.57\%±2.2} & 88.93\%±3.88 & 91.79\%±2.83  \\
        Knuth-Morris-Pratt & 3.91\%±0.15  & 19.51\%±4.57 & 0.03\%±0.1 & 0.01\%±0.02 & \textbf{42.99\%±11.45}  \\ 
        LCS Length & 72.05\%±5.72  & 80.51\%±1.84 & 83.32\%±4.1 & 86.21\%±0.83 & \textbf{86.29\%±1.35}  \\ 
        Matrix Chain Order & 83.91\%±0.49  & 91.68\%±0.59 & 91.89\%±1.2 & \textbf{98.72\%±0.69} & 92.96\%±0.98  \\ 
        Minimum & 87.71\%±0.52  & 97.78\%±0.55 & 95.28\%±2.0 & 95.09\%±8.50 & \textbf{98.50\%±0.10}  \\ 
        MST-Kruskal & 70.97\%±1.50  & 89.80\%±0.77 & 64.91\%±11.8 & 73.41\%±0.98 & \textbf{91.04\%±0.22}  \\ 
        MST-Prim & 69.08\%±7.56  & 86.39\%±1.33 & 85.77\%±7.9 & 89.62\%±2.80 & \textbf{93.08\%±2.66}  \\
        Naive String Matcher & 4.24\%±0.98  & 78.67\%±4.99 & 65.01\%±32.3 & 89.14\%±3.71 & \textbf{98.72\%±1.49}  \\
        Optimal BST & 72.03\%±1.21  & 73.77\%±1.48 & 74.4\%±2.6 & 56.80\%±0.89 & \textbf{79.43\%±2.99}  \\
        Quickselect & 3.66\%±0.42  & 0.47\%±0.25 & \textbf{19.18\%±17.3} & 1.42\%±1.42 & 2.81\%±0.54  \\ 
        Quicksort & 73.10\%±0.67  & 64.64\%±5.12 & 39.42\%±13.2 & 15.20\%±0.60 & \textbf{88.56\%±3.24}  \\ 
        Segments Intersect & 93.53\%±0.88  & 97.64\%±0.09 & 84.94\%±2.6 & 86.96\%±1.52 & \textbf{98.09\%±0.13}  \\ 
        SCC & 32.19\%±9.23  & 43.43\%±3.15 & 28.59\%±15.2 & \textbf{84.08\%±5.70} & 51.93\%±4.37  \\ 
        Task Scheduling & 84.89\%±0.91  & 87.25\%±0.35 & 82.93\%±1.8 & 85.69\%±0.50 & \textbf{89.63\%±0.39}  \\
        Topological Sort & 60.45\%±2.69  & 87.27\%±2.67 & 80.62\%±17.5 & \textbf{98.46\%±1.32} & 98.19\%±1.03  \\ \hline
        Overall Average & 66.04\% & 75.98\% & 66.18\% & 70.42\% & \textbf{82.68\%} \\  \hline
        \textgreater 90\% & 7/30 & 11/30 & 7/30 & 9/30 & \textbf{18/30}  \\ 
        \textgreater 80\% & 11/30 & 17/30 & 15/30 & 18/30 & \textbf{22/30}  \\ 
        \textgreater 60\% & 23/30 & 24/30 & 20/30 & 22/30 & \textbf{25/30}  \\ \hline
    \end{tabular}
    \caption{The test score of CEF-GMPNN and CEF-RT on 30 reasoning tasks within CLRS. The best-performing results in each row are highlighted in bold. The column ``Prior Best'' in the table represents the best results among four approaches in the literature: Memnet~\protect\cite{clrs}, PGN~\protect\cite{clrs}, MPNN~\protect\cite{clrs}, and NPQ~\protect\cite{npq}.}
    \label{tab:results}
\end{table*}

\subsection{Comparison with GNN-based Processors}

In this experiment, we compare our method CEF-GMPNN with Triplet-GMPNN across 30 reasoning tasks within the CLRS dataset. The results are shown in~\cref{fig:mpnn}. 
Bar charts illustrating average scores, with standard deviations denoted by black lines, are presented for each task. Additionally, the tasks are arranged in descending order of improvement magnitude to better showcase trends.

From~\cref{fig:mpnn}, we see that our framework demonstrates enhancement across all 30 reasoning tasks, implying that our framework pushes the Triplet-GMPNN architecture towards an improved state. Notably, in tasks related to sorting and strings, such as bubble sort, quick select, Knuth-Morris-Pratt, and heap sort, the framework significantly improves the performance of Triplet-GMPNN.

\subsection{Comparison with Transformer-based Processors}

In this experiment, we compare our method CEF-RT with RT across 30 reasoning tasks within the CLRS dataset. The results are shown in~\cref{fig:rt}. 

We observe a trend different from the previous experiment. The framework enhances the performance of RT across the majority of CLRS datasets. Specifically, in tasks related to graph optimization, such as strongly connected components (SCC) and depth-first search (DFS), there is a noteworthy increase in test scores of over 30\%. However, for sorting tasks where the original RT approach exhibits poor average scores with very large derivations, our framework leads to a decrease in performance.  This phenomenon indicates that for Transformer-based Processors, our framework steers them toward a more ``extreme'' state.


\subsection{Summarized Results and Analysis}

We summarize the experiential results in~\cref{tab:results}. Due to space limitations, we use column ``Prior Best'' to represent the best results among Memnet~\cite{clrs}, PGN~\cite{clrs}, MPNN~\cite{clrs}, and NPQ~\cite{npq}. Note that~\cite{npq} proposed four variants of the NPQ approach, and for each task, we have chosen the best scores among these variants as reference items. 

In the table, we present the specific test scores for each task, along with summarizing the average scores. Additionally, in line with prior work~\cite{triplet}, we provide the respective proportions of tasks achieving scores greater than 90\%, 80\%, and 60\%. We see the following trends:

\begin{itemize}
    \item Our framework yields state-of-the-art results across the majority of the CLRS benchmark, demonstrating higher average scores and lower deviations compared to prior work. The overall average score of CEF-GMPNN is 82.68\%, which is a substantial improvement over previous results.
    
    \item The results show a broader range of tasks where we attain scores exceeding 60\%, 80\%, and 90\%. Specifically, CEF-GMPNN yields a score of at least 90\% in 18 tasks, 80\% for 22 tasks, and 60\% for 20 tasks.

    \item For both Triplet-GMPNN and RT, our framework enhances their performance in most tasks. However, the specific details of the enhancements differ. For Triplet-GMPNN, the improvements brought by the framework are comprehensive, showing enhancement across all tasks. In contrast, for RT, the framework exhibits more significant improvements in certain tasks compared to Triplet-GMPNN, but it also shows performance degradation in tasks where RT initially performs poorly. These results indicate that our framework pushes Triplet-GMPNN towards an improved state while making RT more ``extreme'' in its performance.
\end{itemize}

\section{Conclusion}\label{sec:conclusion}

The paper studies seq-graph reasoning. We introduce a context-enhanced framework for such tasks and show that it can effectively integrate with various existing approaches. Our experiments reveal that the proposed framework consistently enhances the reasoning capabilities of both GNN-based and Transformer-based architectures, achieving new state-of-the-art results on the CLRS benchmark.

There are several intriguing avenues for future research. One interesting direction involves refining our current framework. As discussed in~\cref{sec:intro}, the outcome sequence may form an entire problem-solving or optimization process, potentially represented by an additional reasoning graph. Therefore, it is worthwhile to explore a framework capable of efficiently capturing such graph structures hidden in the data. Another promising direction could be concentrating on the CLRS benchmark, aiming to design novel architectures that could yield further performance improvements.

\newpage

\section*{Acknowledgements}

This work is supported by the National Key Research Project of China under Grant No. 2023YFA1009402, the National Natural Science Foundation of China (No. 62302166), the Scientific and Technological Innovation 2030 Major Projects under Grant 2018AAA0100902, the Shanghai Science and Technology Commission under Grant No.20511100200, the Dean's Fund of Shanghai Key Laboratory of Trustworthy Computing, ECNU, and the Key Laboratory of Interdisciplinary Research of Computation and Economics (SUFE), Ministry of Education.

\bibliographystyle{named}
\bibliography{ijcai24}

\clearpage
\onecolumn
\appendix
\section{Attention-Based Preprocessor}

In this section, we introduce a QKV attention-based preprocessor. Recall two functions in the preprocessor module:
\begin{equation*}
    \begin{aligned}
        \vS^{(t)} &= \vf_{\enhance}\left( \vL^{(t)}, \vC^{(t)} \right) \\
        \vC^{(t+1)} &= \vf_{\update}\left( \vL^{(t)}, \vC^{(t)} \right)
    \end{aligned}
\end{equation*}

We employ a node-wise (resp. edge-wise) concatenation function for $\vf_{\update}$ and a QKV attention network for $\vf_{\enhance}$. For each node $v\in V$, 

\begin{equation*}
    \begin{aligned}
        q_v =& \vf_{\linear\textendash q}\left(\ell_v^{(t)}\right)\\
        k_v =& \vf_{\linear\textendash k}\left( c_v^{(t)}\right)\\
        \vartheta_v =& \vf_{\linear\textendash \vartheta} \left( c_v^{(t)} \right)\\
        s_v^{(t)} =& \softmax \left(\frac{q_v\cdot k_v^T}{\sqrt{d_{k_v}}}\right) \cdot \vartheta_v  \\
        c_v^{(t+1)} &= \left. \ell_v^{(t)} \;\middle\Vert\; c_v^{(t)} \right.
    \end{aligned}
\end{equation*}

The above is a standard QKV attention mechanism. Function $\vf_{\linear\textendash q}, \vf_{\linear\textendash k}, \vf_{\linear\textendash \vartheta}$ are linear layers and $d_{k_v}$ is the dimension of $k_v$. 
In our empirical evaluation, we apply such an implementation to the Triplet-GMPNN processor and call it CEF-GMPNN-ATT.

It is evident that the demand for memory substantially increases with a rise in the total number of reasoning steps, as $c_v^{(t+1)}$ grows larger over successive steps. This approach proves to be excessively resource-intensive and, thus, is not suitable for all tasks in CLRS, especially those tasks with a large number of reasoning steps like quicksort (2079 steps).
Thus, in the experiments, we pick several tasks with a modest graph size and a limited number of reasoning steps and compare CEF-GMPNN-ATT with CEF-GMPNN on these tasks. The experimental results are presented in~\cref{tab:ablation2}. The results are averaged over three runs with random seeds of 18, 30, and 42. We use $\uparrow$ and $\downarrow$ after each item in that column to represent the increase and decrease in inference performance. From the table, we observe that with an additional memory, CEF-GMPNN-ATT outperforms CEF-GMPNN in certain tasks, particularly in the case of Heapsort.


\begin{table}[htb]
\centering
\begin{tabular}{ccccc}
\hline
\multirow{2}{*}{Tasks} & \multicolumn{2}{c}{Accuarcy} & \multicolumn{2}{c}{Model Size} \\ \cline{2-5} 
                       & CEF-GMPNN   & CEF-GMPNN-ATT  & CEF-GMPNN      & CEF-GMPNN-ATT     \\ \hline
BFS                    & 99.95\%±0.09          & 99.97\%±0.02$\uparrow$             & 1.5 M             & 15 M                \\ \hline
Dijkstra               & 97.89\%±0.22 & 98.20\%±0.71$\uparrow$             & 1.5 M            & 97.5 M               \\ \hline
Graham Scan            & 96.27\%±0.37 & 95.12\%±0.81$\downarrow$       & 1.5 M              & 180 M                 \\ \hline
Heapsort               & 52.13\%±4.91 & 80.55\%±4.93$\uparrow$             & 1.5 M             & 700.5 M                \\ \hline
MST-Prim               & 93.08\%±2.66 & 92.77\%±0.41$\downarrow$           & 1.5 M              & 97.5 M                 \\ \hline
\end{tabular}
\caption{Comparison between CEF-GMPNN and CEF-GMPNN-ATT.}
    \label{tab:ablation2}
\end{table}

\clearpage
\section{Ablation Study of Gating Mechanism Implementation}


In this section, we conduct three ablation experiments on some tasks in CLRS to investigate the effects of certain algorithms. All ablation experiments are repeated three times with random seeds of 18, 30, and 42. Apart from the experimental settings described for each experiment, all other settings remain consistent with those mentioned in the main text. In all the tables, the column containing the ablation experiments is referred to as ``Ablation". In addition, we use $\uparrow$ and $\downarrow$ after each item in that column to represent the increase and decrease in inference performance. 

The first experiment is performed on the CEF-GMPNN model, where the activation function for calculating the forget factor is replaced from $\otanh+\relu$ to $\sigmoid$. The experimental results are presented in \cref{tab:ablation1}. Based on the results of this experiment, we can conclude that when using the CEF framework in GNN-based architectures, the model tends to set the forget factor's value to 0, which can be effectively achieved through the combination of $\otanh+\relu$.

\begin{table}[htb]
    \centering
    \begin{tabular}{ccc}
    \hline
        Tasks & CEF-GMPNN & Ablation  \\ \hline
        Binary Search & 83.91\%±0.70 & 81.36\%±6.14$\downarrow$  \\ 
        Find Max.Subarray & 76.38\%±3.04 & 72.94\%±5.31$\downarrow$  \\ 
        Matrix Chain Order & 92.96\%±0.98 & 90.51\%±0.82$\downarrow$  \\ 
        MST-Prim & 93.08\%±2.66 & 89.70\%±1.94$\downarrow$  \\ 
        Naive String Matcher & 98.72\%±1.49 & 84.51\%±8.95$\downarrow$  \\ \hline
    \end{tabular}
    \caption{Experimental results of using the $\sigmoid$ function to compute the forget factor in CEF-GMPNN.}
    \label{tab:ablation1}
\end{table}

The second experiment replaces the $\sigmoid$ activation function with $\otanh+\relu$ for calculating the forget factor in CEF-RT, which is converse to the first experiment. The experimental results are presented in \cref{tab:ablation3}. It can be observed that the Transformer-based processor, when combined with the CEF framework, prefers the $\sigmoid$ activation function. An intuitive explanation is that such processors require more flexibility in assigning values to the forget factor compared to the GNN-based model architecture.   

\begin{table}[htb]
    \centering
    \begin{tabular}{ccc}
    \hline
        Tasks & CEF-RT & Ablation  \\ \hline
        Bridges & 72.21\%±2.24 & 48.14\%±18.47$\downarrow$  \\ 
        DFS & 75.11\%±10.23 & 73.94\%±15.47$\downarrow$  \\ 
        Find Max.Subarray & 96.63\%±1.77 & 87.05\%±9.42$\downarrow$  \\ 
        SCC & 84.08\%±5.70 & 56.12\%±17.04$\downarrow$  \\ 
        Topological Sort & 98.46\%±1.32 & 99.03\%±0.57$\uparrow$  \\ \hline
    \end{tabular}
    \caption{Experimental results of using the $\otanh+\relu$ function to compute the forget factor in CEF-RT.}
    \label{tab:ablation3}
\end{table}

The third experiment involves replacing the operation of using $C$ as the key and value in the attention mechanism in CEF-RT with the operation of directly assigning $C$ to $z$, similar to CEF-GMPNN. The algorithm for updating node features with attention then utilized the self-attention mechanism from RT. The experimental results are presented in \cref{tab:ablation4}. Through this experiment, we demonstrate the necessity of using a contextual storage module as the key and value for cross-attention in the Transformer architecture within the CEF framework.

\begin{table}[htb]
    \centering
    \begin{tabular}{ccc}
    \hline
        Tasks & CEF-RT & Ablation  \\ \hline
        Bridges & 72.21\%±2.24 & 60.31\%±14.78$\downarrow$  \\ 
        DFS & 75.11\%±10.23 & 75.24\%±9.30$\uparrow$  \\ 
        Find Max.Subarray & 96.63\%±1.77 & 96.32\%±0.87$\downarrow$  \\ 
        SCC & 84.08\%±5.70 & 83.30\%±6.94$\downarrow$  \\ 
        Topological Sort & 98.46\%±1.32 & 98.39\%±1.43$\downarrow$  \\ \hline
    \end{tabular}
    \caption{Experimental results of removing cross attention in CEF-RT.}
    \label{tab:ablation4}
\end{table}

\clearpage
\section{Exploration of the Forget Factor}
In order to further explore the selection of the forget factor in CEF for GNN-based and Transformer-based architectures, we conducted experiments using different values of the forget factor in CEF-GMPNN and CEF-RT. The experimental results were visualized as two heatmaps shown below. From \cref{fig:mpnn_alpha} and \cref{fig:rt_alpha}, it can be observed that CEF-GMPNN performs best when the forget factor value is close to 0, while CEF-RT does not exhibit such a trend.

\begin{figure}[H]
    \centering
    \includegraphics[width=0.55\textwidth]{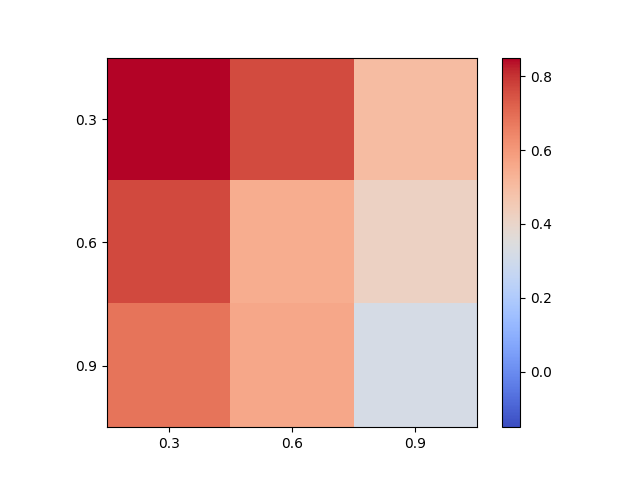}
    \caption{Experimental results of training the HeapSort algorithm on the CEF-GMPNN model using different forget factors. The horizontal axis represents the forgetting rate of storing the $s_v$ context, while the vertical axis represents the forgetting rate of storing the $h_v$ context.}
    \label{fig:mpnn_alpha}
\end{figure}

\begin{figure}[H]
    \centering
    \includegraphics[width=0.55\textwidth]{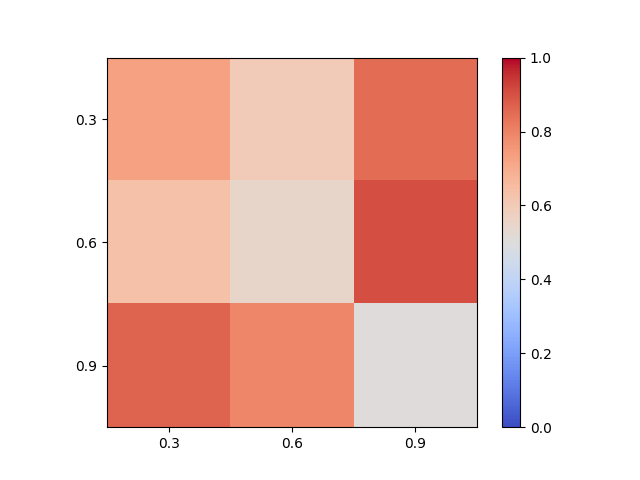}
    \caption{Experimental results of training the Finding Maximum Subarray algorithm on the CEF-RT model using different forget factors. The horizontal axis represents the forgetting rate of storing the context about node features, while the vertical axis represents the forgetting rate of storing the context about edge features.}
    \label{fig:rt_alpha}
\end{figure}

\clearpage
\section{Training Time}\label{sec:time}

In this section, we present a comparative analysis of the training time before and after integrating the CEF framework with Relational Transformer and Triplet-GMPNN. From \cref{tab:time}, it can be observed that there is no significant improvement in training time after incorporating our framework. It serves as evidence for the efficiency of CEF.

\begin{table*}[h]
    \centering
    \begin{tabular}{ccccc}
    \hline
        Tasks & RT & CEF-RT & Triplet-GMPNN & CEF-GMPNN \\ \hline
        Activity Selector & 29min & 30min & 2min & 2min  \\ 
        Articulation Points & 70min & 73min & 28min & 31min  \\
        Bellman-Ford & 25min & 26min & 4min & 4min  \\ 
        BFS & 25min & 26min & 3min & 5min  \\ 
        Binary Search & 33min & 34min & 2min & 2min  \\ 
        Bridges & 70min & 73min & 30min & 32min  \\ 
        Bubble Sort & 80min & 83min & 28min & 32min  \\ 
        DAG Shortest Paths & 52min & 54min & 12min & 13min  \\
        DFS & 50min & 53min & 12min & 17min  \\ 
        Dijkstra & 31min & 33min & 4min & 4min  \\ 
        Find Max.Subarray & 49min & 52min & 3min & 3min  \\ 
        Floyd-Warshall & 39min & 40min & 10min & 10min  \\ 
        Graham Scan & 39min & 41min & 6min & 6min  \\
        Heapsort & 70min & 73min & 31min & 37min  \\ 
        Insertion Sort & 29min & 29min & 6min & 6min  \\ 
        Jarvis’March & 122min & 129min & 18min & 19min  \\ 
        Knuth-Morris-Pratt & 113min & 121min & 14min & 16min  \\ 
        LCS Length & 30min & 31min & 4min & 4min  \\ 
        Matrix Chain Order & 39min & 40min & 13min & 13min  \\ 
        Minimum & 59min & 61min & 3min & 3min  \\ 
        MST-Kruskal & 51min & 53min & 12min & 13min  \\ 
        MST-Prim & 31min & 32min & 3min & 3min  \\ 
        Naive String Matcher & 93min & 97min & 7min & 7min  \\ 
        Optimal BST & 43min & 44min & 8min & 8min  \\ 
        Quickselect & 266min & 278min & 24min & 26min  \\ 
        Quicksort & 72min & 74min & 30min & 32min  \\ 
        Segments Intersect & 33min & 34min & 2min & 2min  \\ 
        SCC & 79min & 81min & 25min & 26min  \\ 
        Task Scheduling & 31min & 31min & 2min & 2min  \\ 
        Topological Sort & 46min & 48min & 11min & 11min  \\ \hline
        Overall Average  & 60.0min & 62.5min & 11.9min & 13.0min \\ \hline
    \end{tabular}
    \caption{A comparative analysis of the training time before and after integrating the CEF framework for Relational Transformer and Triplet-GMPNN models is presented.}
    \label{tab:time}
\end{table*}





\end{document}